%% file: instate.tex
\documentclass[11pt,  letterpaper]{article}

\usepackage{mathpazo}

\usepackage{csvsimple}

\usepackage{booktabs}

\input{boilerplate}

\title{Instate: Predicting the State of Residence From Last Name

\thanks{Data and scripts needed to replicate the results are available at: https://github.com/appeler/instate/}}

\author{Atul Dhingra\thanks{Senior Member, IEEE; \textsf{dhingra.atul92@gmail.com}} \and Gaurav Sood\thanks{Independent researcher;  \textsf{gsood07@gmail.com}.}}

\date{\today}

\begin{document}

\maketitle

\begin{abstract}

India has twenty-two official languages. Serving such a diverse language base is a challenge for survey statisticians, call center operators, software developers, and other such service providers. To help provide better services to different language communities via better localization, we introduce a new machine learning model that predicts the language(s) that the user can speak from their name. Using nearly 438M records spanning 33 Indian states and 1.13M unique last names from the Indian Electoral Rolls Corpus \citep{DVN/OG47IV_2023}, we build a character-level transformer-based machine-learning model that predicts the state of residence based on the last name. The model has a top-3 accuracy of 85.3\% on unseen names. We map the states to languages using the Indian census to infer languages understood by the respondent. We provide open-source software that implements the method discussed in the paper.

\smallskip



\end{abstract}








\section{Introduction}

An overwhelming majority of people in India don’t speak the language in which the vast majority of web content and services are delivered. 57\% of the web content is in English.\footnote{According to this \href{https://w3techs.com/technologies/overview/content_language}{article}, English is used by 57\% of websites whose language is decipherable.} And according to the last Indian census, less than 150M Indians of the more than 1.2B could speak English (and likely much fewer can read it). Hence, localization is essential for delivering software and services. Localization proceeds in two steps: inferring (or letting users select) the language understood by the user and rendering the content or service in that language. In this paper, we present a way to improve localization in India. In particular, we provide a machine-learning model that predicts the languages that are known to the user. To infer the language(s) understood by the user, we rely on names. Using nearly 438M records spanning 33 states and 1.13M unique last names from the Indian Electoral Rolls Corpus \citep{DVN/OG47IV_2023}, we build a character-level machine-learning model that predicts the state of residence based on the last name. The best-performing model has a top-3 accuracy of 85.3\%  on unseen names. Using the data from the Indian census, we map the states to languages and provide a prediction for the languages understood by the respondent. We also provide \href{https://github.com/in-rolls/instate}{open-source software} that implements the method discussed in the paper.

\section{Related Work}
Learning sociodemographic characteristics of people, e.g., race and ethnicity, from their names is a well-studied problem \citep{imai2016improving, parasurama2021racebert, wong2020machine, ye2017nationality, sood2018predicting}. A variety of techniques have been invented for this purpose. For instance, \citet{blevins2015jane} use a Naive Bayes classifier on historical census data. \citet{imai2016improving} combine voter registration data with census estimates to infer the race/ethnicity of a person given their last name. Other researchers use embeddings of email and other public social networks to infer ethnicity \citep{ye2017nationality, hur2022malaysian, junting_skiena}. Yet others have used deep learning approaches to exploit the patterns of letters in the name to infer sociodemographic characteristics \citep{hu2021s, sood2018predicting, parasurama2021racebert}. Our paper builds on this extensive body of work to infer what potential languages a person may be able to speak.








\section{Data}

We exploit the Indian electoral rolls data \citep{DVN/OG47IV_2023, DVN/MUEGDT_2018} to build the machine learning model. The corpus includes data on nearly 438M people from 33 states with nearly 1.14 million unique last name spellings. The electoral roll data includes information on the elector's husband or father’s name, age, sex, house number, and geographical details, like the polling station, constituency, local police station, etc. Electoral rolls in India are a particularly useful source of data given that they are a proximate census of the adult population. Unlike the US, in India, a neutral administrative body called the Election Commission works to produce an exhaustive list of eligible voters—all adult citizens. The electoral rolls are published online ahead of the elections to make sure that no one is left out, and so that others can point to any errors, e.g., dead voters, missing voters, etc. There are, however, some problems with the data. Primarily, voters are registered in their `home' districts, even if they are working elsewhere. This works to our advantage to the extent that we rely on name, and state mappings to infer the languages that the user can understand.

The parsed electoral rolls corpus \citep{DVN/MUEGDT_2018} includes transliterations to English using the Python package indicate \citep{Chintalapati_Indicate_Transliterate_Indic_2022}. There is no unique English transliteration of Hindi names and the package provides the most probable transliteration. 

We start by collating the data, preserving only five columns: first name, last name, father's or husband's name, sex, and state. To establish the last name of each person, we work as follows: 1. if the person's name is more than one word, we assume the last word to be the person's last name, 2. else, we check if the father's or husband's name is more than one word and where it is, we assume the last word to be the last name. We discard all other rows as cases where we couldn't establish the last name. We further discard cases where the last name is less than three letters as manual analysis of the data suggests that these are not last names. We also remove last names that have non-alphanumeric characters. Finally, to preserve privacy and because really infrequent last names are unlikely to be true last names, we exclude last names that occur less than thrice across the dataset. Finally, we convert all the last names into lowercase. 

\section{Models}

If we have no information from the electoral rolls, then given the last name, the optimal prediction for the state of residence is the most populous state. In our case, that is Uttar Pradesh, which has nearly 230M people. If we did that, we would be right about 16.74\% of the time.

If we assume that electoral rolls constitute the universe of people in India with the correct, unique transliterated English spellings of their names, and if the last name is the only piece of information we have about the person, the Bayes Optimal classifier is simply an intercept-only model that gives the population mean---the proportion of last names in various states. 

The Naive Bayes model rests on untenable assumptions. It assumes no data entry errors, unique transliterations, and a universe of the adult population. (We know that we have only a fraction of the total adult population of India, which is nearly a billion.) 


To learn a more generalizable representation that rests on more tenable assumptions, we test three deep learning models---RNN, LSTM, and GRU. The setup for all three models is similar. We split the data by last names and keep 80\% of the data for training and 20\% for testing. For the test data, we once again group by last names so that we have as many rows as unique names in the test set. For the training data, we transform the data to preserve all the unique names per state, which leaves us with approximately 1.5M rows. For each of the models, we predict the top-3 states and incur a loss if our predictions don't include any of the top-3 states. Before describing the results, we discuss the motivation and setup for each of the models:

\begin{itemize}

    \item \textbf{RNN} We estimate a 2-layer RNN \citep{10.1145/3426826.3426842, 8469258} with 512 hidden units, train the model with a batch size of 256 using SGD optimizer and use a negative log-likelihood loss. To deal with exploding gradients, we set a low learning rate of .005 and a momentum of .9.

    \item \textbf{LSTM} While training the RNN we observed that the training loss was slow to converge. To increase the model capacity while avoiding common issues with deep RNN models---exploding or vanishing gradients---we build an LSTM model \citep{hochreiter1997long}. We estimate a 2-layer LSTM with 512 hidden units, train the model with a batch size of 256 using the Adam optimizer and use a negative log-likelihood loss. We set a low learning rate of 3e-4. 

    \item \textbf{GRU} One of the drawbacks of using an LSTM is that it is very slow to train and converge. Therefore, to add more capacity to the model, we chose GRU which we could train for longer and converged faster. We estimated a 2-layer GRU \citep{cho2014properties} with 2048 hidden units and trained the model with a batch size of 1024 using the Adam optimizer with a learning rate of 3e-4 and a negative log-likelihood loss. 

\end{itemize}

We comprehensively evaluate the performance of these models. We compare the performance of the models on the test set, a weighted random sample of 3000 unique names with weights proportional to popularity, 3000 most popular names, and 3000 least popular names. Table \ref{model_comparison} summarizes the results. As the table shows, GRU is the best-performing model across the board. The best-performing GRU model has a top-3 accuracy of 85.3\% in the test set. Across various slices, the accuracy of the GRU model never dips below 82\%. 

To complement these results, we also calculated the more conventional accuracy metric: the probability that our top prediction is the modal state in which the person with the last name lives. The accuracy for RNN, LSTM, and GRU is 47.6\%, 34.4\%, and 57.9\% respectively.

\input{tabs/model_comparison}

\begin{figure}[!htb]

  \centering

    \caption{Relationship Between Popularity and Accuracy Across Various Deep Learning Models}

  \includegraphics[]{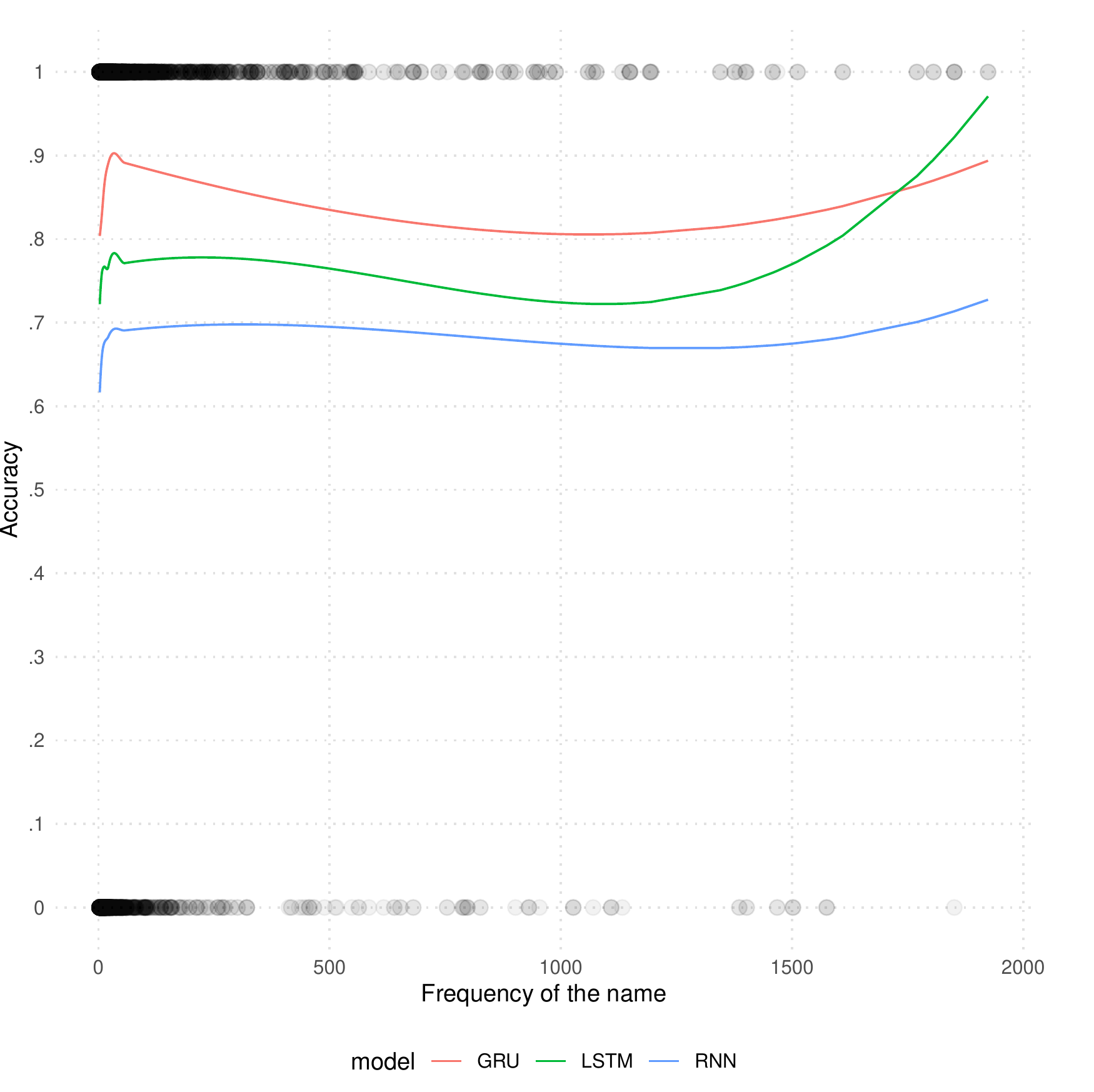}

  \label{fig:popularity_accuracy}

\end{figure}

To dig deeper into the relationship between accuracy and popularity, we plot a lowess. Given the extreme skew in the distribution of last names, we truncate the sample to last names shared by at most 2000 people. As Figure \ref{fig:popularity_accuracy} shows, GRU wins handily except at the tail end where we have very limited data.

\begin{figure}[!htb]

  \centering

    \caption{Relationship Between Gender Ratio and Accuracy Across Various Deep Learning Models}

  \includegraphics[]{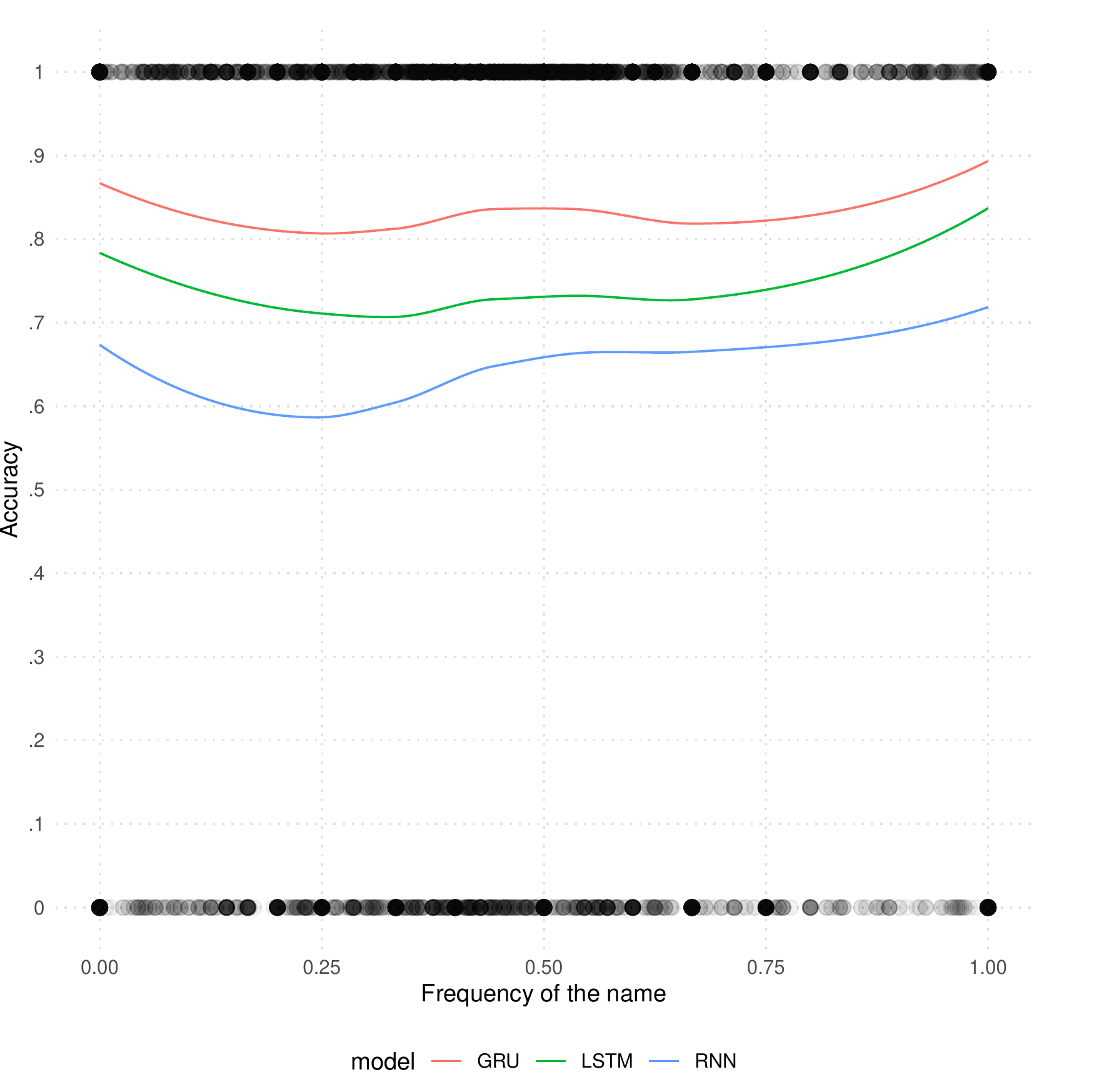}

  \label{fig:women_accuracy}

\end{figure}

We also investigate how accuracy varies by the proportion of women with the last name. To investigate, we once again plot a lowess. Figure \ref{fig:women_accuracy} makes clear two patterns. First, the lines are almost parallel to the x-axis, suggesting consistent performance across the range. Second, GRU has a consistent advantage over LSTM which in turn has a consistent advantage over RNN.

Lastly, we check how the models perform across states. We take a random sample of 1000 unique names per state and estimate accuracy within each state. Table \ref{accuracy_by_state} summarizes the results. GRU once again has a clear performance advantage in most states and runs close to the best-performing model in states where it isn't the outright winner, e.g., Andaman and Nicobar, Arunachal Pradesh, etc.

\input{tabs/accuracy_by_state}

\section{States to Languages}

To create our states to languages database, we consulted the state constitutions for the official languages and the latest Indian census (the 2011 census) for the most widely spoken languages in the state. We then append the mapping to our state predictions. Our method has some weaknesses. While the modern Indian state boundaries and state education curricula are affected by language conflicts (for instance, there were \href{https://en.wikipedia.org/wiki/Anti-Hindi_agitations_of_Tamil_Nadu}{anti-Hindi agitations in Tamil Nadu}), there is no unique mapping between states and languages. This is for a few reasons. For one, many Indian states are massive and there is considerable linguistic diversity across regions within states. Second, internal migration, especially to urban centers, has transformed many cities. For instance, say that all Dhingras only speak Punjabi and say that all of them once upon a time lived in Punjab. Suppose that over time half of them immigrated to Delhi and never learned another language. Say that we map Delhi to people who understand Hindi. Our classifier will then indicate that Dhingras have a 50\% chance of speaking Punjabi and a 50\% chance of speaking Hindi. But in fact, no Dhingra can understand Hindi. Our prediction in such a case would be highly erroneous. We have reasons to think, however, that the net error is generally likely lower. For instance, in the above example, we make the slightly untenable assumption that immigrants never learn to understand the dominant language of the geography they immigrate to. This can be true but it is also likely that there is both selection bias—Dhingras who are open to learning new languages or may already know another language are likelier to immigrate to Delhi—and learning—Dhingras who immigrate to Delhi learn a new language over time. The other compensating virtue of our dataset, as we noted above, is that electoral rolls still map immigrants to their home districts. To the extent that immigrants do not forget their mother tongue, we will have good predictions. 

\section{Discussion}

Our paper contributes to technologies that improve localization. However, our method of predicting languages understood by the user from their last name has a few limitations. The first is that our geographical units are very crude. Indian states are often as big as countries. For instance, Uttar Pradesh has more than 230 million people. There is also a large linguistic diversity within the states. In future versions, we hope to exploit the more granular data provided in the electoral rolls. Second, the relationship between geography and language is weaker because of internal migration. This is especially true in urban areas. Though as we note, our unique database of electoral rolls has the virtue of mapping people to their home districts. The third challenge is transliteration. There is no one standard way of transliterating Indian languages into English. We use one transliteration when the way to improve coverage may be to use k-best transliterations with the transliterations produced roughly in the frequency of how common they are. The fourth challenge is establishing last names. Our method of establishing the last name is crude. One way to improve the heuristic is to only use last words in names that are common across household members’ names. 

\newpage

\bibliography{instate.bib}

\end{document}

%% file: boilerplate.tex
\usepackage{amsfonts}

\usepackage{pstricks,fullpage, subfigure, 
  epsfig,amsthm,amsmath,latexsym,amssymb,verbatim,url,setspace,multirow,fancyhdr,
  pdfsync,rotating,xfrac,footnote,pst-plot}

\usepackage{graphicx}
\usepackage{tikz}
\usetikzlibrary{matrix}
\usepackage[bottom]{footmisc}
\usepackage[round,comma]{natbib}
\usepackage{rotating}
\usepackage[bottom]{footmisc}
\usepackage{bbm}
\usepackage[font=footnotesize,labelfont=bf]{caption}

\usepackage{lscape}
\newcommand{\bland}{\begin{landscape}}
\newcommand{\eland}{\end{landscape}}

\usepackage{hyperref}

\hypersetup{
   colorlinks=true,           linkcolor=black,
   citecolor=black,            filecolor=magenta,
   urlcolor=darkgray}

\newcommand{\noi}{\noindent}

\newcounter{Lcount}

\graphicspath{{figs/}}

\usepackage[
    linecolor=red,
    bordercolor=red,
    backgroundcolor=white,
    textsize=tiny
]{todonotes}

\usepackage{
  amscd,
  amsfonts,
  amsmath,
  amssymb,
  amsbsy,
  amsthm,
  bm, 
  dsfont,
  cancel, 
  latexsym,
  cases,
  mathtools,
  graphicx,
  xcolor,
  xargs,
  lineno
}

\usepackage{algorithm2e}
\SetKwInput{kwParam}{Parameters}

\newcommand{\beq}{\begin{equation}}
\newcommand{\eeq}{\end{equation}}

\newcommand*\Bigpar[1]{\left( #1 \right )}

\newcommand{\ba}{\begin{array}}
\newcommand{\ea}{\end{array}}
\newcommand{\be}{\begin{enumerate}}
\newcommand{\ee}{\end{enumerate}}
\newcommand{\bi}{\begin{itemize}}
\newcommand{\ei}{\end{itemize}}
\newcommand{\bs}{\begin{align}\begin{split}\nonumber}
\newcommand{\bsnumber}{\begin{align}\begin{split}}
\newcommand{\es}{\end{split}\end{align}}

\newcommandx{\deriv}[2][1=x,2=f]{\nabla \, #2 \Bigpar{ #1 } }
\newcommandx{\ortho}[1][1=L]{#1^{\bot}}

\newcommandx*\seqq[3][1=1,2=x, 3=n]{#2_{#1},\ldots,#2_{#3}}
\newcommandx*\coord[3][1=1,2=x, 3=n]{(#2_{#1},\ldots,#2_{#3})}

\newcommand\frakfamily{\usefont{U}{yfrak}{m}{n}}
\DeclareTextFontCommand{\textfrak}{\frakfamily}

\newcommand{\hyp}[2]{
\ensuremath{H_0:#1 \ifhmode\quad\text{versus}\quad\fi\text{ vs. } H_1:#2}}

\newcommandx{\uniff}[1][1={a,b}]{\textrm{Unif}\left({#1}\right)}
\newcommandx{\unifd}[1][1={a,\ldots,b}]{\textrm{Unif}\left\{{#1}\right\}}
\newcommandx{\dunif}[3][1=x,2=a,3=b]{\frac{I(#2<#1<#3)}{#3-#2}}
\newcommandx{\dunifd}[3][1=x,2=a,3=b]{\frac{I(#2\le#1\le#3)}{#3-#2+1}}
\newcommandx{\punif}[3][1=x,2=a,3=b]{
\begin{cases} 0 & #1 < #2 \\ \frac{#1-#2}{#3-#2} & #2 < #1 < #3 \\ 1 & #1 > #3\\\end{cases}}
\newcommandx{\punifd}[3][1=x,2=a,3=b]{
\begin{cases} 0 & #1 < #2\\ \frac{\lfloor#1\rfloor-#2+1}{#3-#2} & #2 \le #1 \le #3 \\ 1 & #1 > #3\\ \end{cases}}

\newcommandx\bern[1][1=p]{\textrm{Bern}\left({#1}\right)}
\newcommandx\dbern[2][1=x,2=p]{#2^{#1} \left(1-#2\right)^{1-#1}}
\newcommandx\pbern[2][1=x,2=p]{\left(1-#2\right)^{1-#1}}

\newcommandx\bin[1][1={n,p}]{\textrm{Bin}\left(#1\right)}
\newcommandx\dbin[3][1=x,2=n,3=p]{\binom{#2}{#1}#3^#1\left(1-#3\right)^{#2-#1}}

\newcommandx\mult[1][1={n,p}]{\textrm{Mult}\left(#1\right)}
\newcommandx\dmult[3][1=x,2=n,3=p]{\frac{#2!}{#1_1!\ldots#1_k!}#3_1^{#1_1}\cdots#3_k^{#1_k}}

\newcommandx\hyper[1][1={N,m,n}]{\textrm{Hyp}\left({#1}\right)}
\newcommandx\dhyper[4][1=x,2=N,3=m,4=n]{\frac{\binom{#3}{#1}\binom{#2-#3}{#4-#1}}{\binom{#2}{#4}}}

\newcommandx\nbin[1][1={r,p}]{\textrm{NBin}\left({#1}\right)}
\newcommandx\dnbin[3][1=x,2=r,3=p]{\binom{#1+#2-1}{#2-1}#3^#2(1-#3)^#1}
\newcommandx\pnbin[3][1=x,2=r,3=p]{I_#3(#2,#1+1)}

\newcommandx\geo[1][1=p]{\textrm{Geo}\left(#1\right)}
\newcommandx\dgeo[2][1=x,2=p]{#2(1-#2)^{#1-1}}
\newcommandx\pgeo[2][1=x,2=p]{1-(1-#2)^#1}

\newcommandx\pois[1][1=\lambda]{\textrm{Po}\left({#1}\right)}
\newcommandx\dpois[2][1=x,2=\lambda]{\frac{#2^#1 e^{-#2}}{#1!}}
\newcommandx\ppois[2][1=x,2=\lambda]{e^{-#2}\sum_{i=0}^#1\frac{#2^i}{i!}}

\newcommandx\normall[1][1={\mu,\sigma^2}]{\mathcal{N}\left({#1}\right)}
\newcommandx\dnormall[3][1=x,2=\mu,3=\sigma]%
  {\frac{1}{#3\sqrt{2\pi}}\exp \Bigpar{-\frac{\left(#1-#2\right)^2}{2 #3^2}}}
\newcommandx\pnormall[1][1=x]{\Phi\left({#1}\right)}
\newcommandx\qnormall[1]{\Phi^{-1}\left({#1}\right)}

\newcommandx\mvn[1][1={\mu,\Sigma}]{\mathrm{MVN}\left({#1}\right)}

\newcommandx\ex[1][1=\lambda]{\textrm{Exp}\left(#1\right)}
\newcommandx\dex[2][1=x,2=\lambda]{#2e^{-#1 #2}}
\newcommandx\pex[2][1=x,2=\lambda]{1-e^{-#1 #2}}

\newcommandx\gam[1][1={\alpha,\lambda}]{\textrm{Gamma}\left({#1}\right)}
\newcommandx\dgamma[3][1=x,2=\alpha,3=\lambda]%
  {\frac{#3^{#2}}{\Gamma\left( #2 \right)} #1^{#2-1}e^{-#3#1}}

\newcommandx\invgamma[1][1={\alpha,\beta}]{\textrm{InvGamma}\left({#1}\right)}
\newcommandx\dinvgamma[3][1=x,2=\alpha,3=\beta]%
{\frac{#3^{#2}}{\Gamma\left(#2\right)}#1^{-#2-1}e^{-#3/#1}}
\newcommandx\pinvgamma[3][1=x,2=\alpha,3=\beta]%
{\frac{\Gamma\left(#2,\frac{#3}{#1}\right)}{\Gamma\left(#2\right)}}

\newcommandx\bet[1][1={\alpha,\beta}]{\textrm{Beta}\left(#1\right)}
\newcommandx\dbeta[3][1=x,2=\alpha,3=\beta]
{\frac{\Gamma\left(#2+#3\right)}{\Gamma\left(#2\right)\Gamma\left(#3\right)}#1^{#2-1}\left(1-#1\right)^{#3-1}}

\newcommandx\dir[1][1={\alpha}]{\textrm{Dir}\left(#1\right)}
\newcommandx\ddir[3][1=x,2=\alpha]{\frac{\Gamma\left(\sum_{i=1}^k #2_i\right)}{\prod_{i=1}^k\Gamma\left(#2_i\right)}\prod_{i=1}^k #1_i^{#2_i-1}}

\newcommandx\weibull[1][1={\alpha}]{\textrm{Dir}\left(#1\right)}
\newcommandx\dweibull[3][1=x,2=\lambda,3=k]{\frac{#3}{#2}
\left(\frac{#1}{#2}\right)^{#3-1} e^{-(#1/#2)^k}}

\newcommandx\chisq[1][1=k]{\chi_{#1}^2}

%% file: tabs/model_comparison.tex
\begin{table}
\centering
\caption{Accuracy of Different Models on the Test Set}
\label{model_comparison}
\begin{tabular}{lrrrr}
\toprule
Model & Test Set & Weighted Random (3k) & Top-3k & Bottom-3k \\
\midrule
RNN & 65.3 & 65.4 & 62.7 & 62.3 \\
LSTM & 75.9 & 75.4 & 73.1 & 72.4 \\
GRU & 85.3 & 84.1 & 82.4 & 82.0 \\
\bottomrule
\end{tabular}
\end{table}

%% file: tabs/accuracy_by_state.tex
\begin{table}
\centering
\caption{Accuracy of Different Models By State of Residence}
\label{accuracy_by_state}
\begin{tabular}{lrrr}
\toprule
state & RNN & LSTM & GRU \\
\midrule
Andaman and Nicobar & 50.2 & 69.2 & 66.9 \\
Andhra Pradesh & 65.2 & 70.1 & 80.4 \\
Arunachal Pradesh & 60.0 & 81.7 & 81.2 \\
Assam & 73.4 & 93.6 & 89.0 \\
Bihar & 25.7 & 36.5 & 90.2 \\
Chandigarh & 18.6 & 23.3 & 88.4 \\
Dadra & 69.7 & 76.9 & 79.7 \\
Daman & 44.8 & 50.9 & 61.1 \\
Delhi & 22.4 & 36.2 & 37.6 \\
Goa & 34.5 & 48.1 & 53.5 \\
Gujarat & 88.1 & 91.6 & 94.7 \\
Haryana & 18.2 & 18.7 & 94.3 \\
Jharkhand & 32.2 & 45.3 & 79.1 \\
Jammu and Kashmir & 68.8 & 84.9 & 89.1 \\
Karnataka & 88.5 & 89.8 & 94.4 \\
Kerala & 26.5 & 52.4 & 50.7 \\
Maharashtra & 50.0 & 67.7 & 72.3 \\
Manipur & 31.0 & 49.8 & 54.3 \\
Meghalaya & 27.8 & 93.6 & 87.7 \\
Mizoram & 79.7 & 85.4 & 86.0 \\
Madhya Pradesh & 23.5 & 29.1 & 75.9 \\
Nagaland & 61.1 & 77.7 & 82.7 \\
Odisha & 76.0 & 91.0 & 89.4 \\
Puducherry & 36.2 & 37.5 & 54.3 \\
Punjab & 13.3 & 16.3 & 97.1 \\
Rajasthan & 14.3 & 15.6 & 81.9 \\
Sikkim & 71.3 & 96.3 & 90.2 \\
Telengana & 98.0 & 96.5 & 96.8 \\
Tripura & 88.3 & 99.2 & 97.0 \\
Uttar Pradesh & 12.6 & 18.0 & 86.6 \\
Uttaranchal & 16.6 & 21.2 & 80.1 \\
\bottomrule
\end{tabular}
\end{table}